\documentclass[letterpaper, 10 pt]{ieeeconf}
\IEEEoverridecommandlockouts 
\usepackage{cite}
\usepackage{bbm}
\usepackage{mathtools}
\usepackage{gensymb}
\usepackage{tabu}
\usepackage[bf]{caption}
\usepackage{amsmath}
\usepackage{algorithm}
\usepackage[noend]{algpseudocode}
\usepackage{breqn}
\usepackage{subcaption}
\usepackage{caption}
\usepackage{float}
\usepackage{float}

\makeatletter
\def\BState{\State\hskip-\ALG@thistlm}
\makeatother

\title{\LARGE \bf
Avoidance of Manual Labeling in Robotic Autonomous Navigation Through Multi-Sensory Semi-Supervised Learning on Big Data
}

\author{Junhong Xu, Shangyue Zhu, Hanqing Guo and Shaoen Wu% <-this % stops a space
\thanks{Authors are with the Department of Computer Science, Ball State University, Munice, Indiana.
        {\tt\small \{jxu7,szhu,hguo,swu\}@bsu.edu}}%
}

\begin{document}
% \title{Avoidance of Manual Labeling in Robotic Autonomous Navigation Through Multi-Sensory Semi-Supervised Learning}
%\author
%{\IEEEauthorblockN{Junhong Xu, Shangyue Zhu, Hanqing Guo, Shaoen Wu}
%\IEEEauthorblockA{Computer Science\\ Ball State University\\}
%}

\maketitle

\begin{abstract}
Imitation learning holds the promise to address challenging robotic tasks such as autonomous navigation. It however requires a human supervisor to oversee the training process and send correct control commands to robots without feedback, which is always prone to error and expensive. To minimize human involvement and avoid manual labeling of data in the robotic autonomous navigation with imitation learning, this paper proposes a novel semi-supervised imitation learning solution based on a multi-sensory design. This solution includes a suboptimal {\it sensor policy} based on sensor fusion to automatically label states encountered by a robot to avoid human supervision during training. In addition, a {\it recording policy} is developed to throttle the adversarial affect of learning too much from the suboptimal sensor policy. This solution allows the robot to learn a navigation policy in a self-supervised manner. With extensive experiments in indoor environments, this solution can achieve near human performance in most of the tasks and even surpasses human performance in case of unexpected events such as hardware failures or human operation errors. To best of our knowledge, this is the first work that synthesizes sensor fusion and imitation learning to enable robotic autonomous navigation in the real world without human supervision. 
\end{abstract}

\section{Introduction} \label{intro}
Indoor mobile robot navigation has a long research history in robotics. Many solutions have been proposed for indoor robotic navigation. One of the most widely adopted approaches is SLAM, which is a two-stage approach consisting of perception and action stage \cite{thrun2008simultaneous}. A global map is built using on-board sensors in the action stage. This map is then given to a motion planner to make predictions \cite{rimon1992exact, tedrake2010lqr}. This approach is environment-dependent and requires tremendous efforts to establish the environment maps. It also needs intensive computation to support the map based prediction.

Deep reinforcement learning (DRL) \cite{mnih2015human} is another attempted exploration to address autonomous indoor navigation. Although it has yielded many successful outcomes \cite{mnih2016asynchronous,Lillicrap2015Continuous,schaul2015prioritized,mirowski2016learning}, it requires robots to learn from trial-and-error that is impractical and too expensive in real world tasks. Therefore, many works only employ DRL for navigation tasks in simulated settings \cite{isele2017navigating} to avoid hardware damaging. The learned policy is then transferred to real world scenarios \cite{tai2017virtual,sadeghi2016cad}. However, the discrepancy between simulated environments and the real world is large and still requires fine-tuning. Meanwhile it is very likely to cause damage to robots. 

Recently, end-to-end imitation learning methods have been employed to resolve complex robotic tasks such as manipulating objects \cite{rahmatizadeh2016learning}, navigating to a target position \cite{pfeiffer2016perception}, and self-driving vehicles \cite{bojarski2016end}. Imitation learning converts sequential decision tasks to supervised learning problems, where a policy is trained to minimize the errors between the predicted actions and the actions taken by an expert policy. However, there are two major challenges when applying this method to sequential decision tasks: (1) data mismatch between states encountered by a trained policy and an expert policy, and (2) difficulties of querying an expert policy in real-world scenarios. The first problem is normally tackled by literature solutions with a DAgger algorithm \cite{ross2011reduction} that iteratively collects training examples using both an expert and a trained policy. This approach however requires a human supervisor to provide correct control commands given an observation encountered by a trained policy, which is expensive and inaccurate \cite{ross2013learning}. Some works have attempted to address the above imitation learning problems by using a hierarchy of supervisors \cite{Laskey2016Robot} or reducing the number of queries to the expert policy \cite{zhang2016query,laskey2016shiv}, 

In this work, we focus on autonomous robotic navigation, we aim at the challenges of using imitation learning in real-world robotic domains where querying a human supervisor is expensive and inaccurate. We propose a solution, \textit{\textbf{M}ulti-\textbf{S}ensory \textbf{S}emi-\textbf{S}upervised \textbf{L}earning (MS3L)}, which is based on imitation learning augmented with sensor fusion. Our solution has two key innovations:
\begin{itemize}
\item One is to employ various types of sensors, including both imaging and non-imaging, in a deep learning framework to minimize human involvement in that it only requires ONE iteration of human supervision to initialize a navigation policy. 
\item The other is that, after initializing the navigation policy, {\it MS3L} uses a suboptimal \textit{sensor policy} to label the observations encountered by a mobile robot at its own, which completely eliminates the need of querying an expert policy in literature solutions. To reduce the adversarial effect of learning from the suboptimal policies, we design a \textit{recording policy} based on the safety policy \cite{zhang2016query}, which controls the degree of information learned by the navigation policy. 
\end{itemize}

In the rest of this paper, Section \ref{sec:related} reviews the related work, particularly imitation learning and reinforcement learning. Then, Section \ref{sec:design} discusses the design of our proposed solution. Next, Section~\ref{sec:eval} presents the extensive performance evaluations of {\it MS3L} in real indoor environments. The paper is finally concluded by Section~\ref{sec:conclusion}.

\section{Related Work}\label{sec:related}
There are a large number of works that address mobile robotic navigation with various methodologies including robotic controls, machine learning, and reinforcement learning. Though, we present the most related works and formulate the research problems in this section.

\subsection{Imitation Learning} 
Imitation learning solutions in literature are unanimously based on a pioneer DAgger algorithm proposed by Ross {\it et al}. \cite{ross2011reduction} to iteratively train a policy that imitates a certain expert policy. This algorithm has been widely used in many robotic problems \cite{ross2013learning,Gupta2017Cognitive,Laskey2016Robot}. These methods require some expert policies to be queried, which is impractical and too expensive in real environments especially when human supervision is required. Ross {\it et al}. use DAgger to train a drone to fly in forests and avoid collisions \cite{ross2013learning}. The human supervisors are provided with partial feedback to correct actions of trained policy offline. While this solution aims at reducing supervisor's burden, it still does not solve the problem of querying a human operator. Laskey {\it et al} proposes an approach to use a hierarchy of supervisors to learn grasping policy \cite{Laskey2016Robot}, which actually requires more on the burden of human supervisor. There are many works extending DAgger algorithm and focusing on the improvement of query efficiency to an expert policy \cite{zhang2016query,laskey2016shiv}. These works are the most relevant to ours in a way that they constrain training data with some query metrics. These methods however assume that an oracle is always available and easy to be queried, which is often unlikely in real environments. In addition, the issue of noisy sensor measurements is not addressed in these works, which definitely degrade the performance of a trained policy.  

Another approach to robotic navigation using imitation learning is to set up multiple cameras to capture training samples in different directions \cite{giusti2016machine,bojarski2016end}. Multiple cameras are installed on a drone or car to collect training samples. Each sample is labeled according to camera positions. This method tackles the data mismatch problem by training the policy with samples from different view directions. However, this data collection strategy only works properly in the domain of lane following. In addition, installing multiple cameras on small robots is challenging or impractical.

\subsection{Reinforcement Learning}
Recently, deep reinforcement learning (DRL) \cite{mnih2015human} has been attracted a lot attention in the robotic control field. Many works based on DRL have been performed to address robotic navigation tasks. In \cite{isele2017navigating}, the authors train a DQN (deep q learning) agent \cite{mnih2015human} to cross an intersection in simulation. Another work proposes a simulated environment to train a DRL agent to reach a target position in indoor environments \cite{Zhu2016Target}. Lillicrap et al. \cite{Lillicrap2015Continuous} proposes deep deterministic policy gradient (DDPG) to train an agent to avoid dynamic obstacles in simulation. These methods however are all constrained in simulated settings where damage to the agents is not a concern. Applying DRL to real environments is still challenging. Many researchers attempt to address this problem by transferring learned DRL policies from simulation to real world navigation tasks \cite{Tobin2017Domain,sadeghi2016cad}. The difference between simulated environments and the real world settings makes this adaption difficult.

\subsection{Learn from Noisy Labels}
Our work is also closely related to the idea of learning from noisy labels in deep neural networks, where a network is trained using a dataset consisting of inaccurate labels. Most works address this problem by modeling the label noise distribution with either a neural network layer or probabilistic graphical models \cite{sukhbaatar2014training, xiao2015learning}. Our work constrains learning from noisy labels by restricting to learn from sub-optimal sensor policy using a recording policy. % Rather than modeling noise distribution, we model the discrepancy between an expert policy and a suboptimal policy that the navigation policy is learned from. This strategy discards the samples labeled by suboptimal policy that deviate too much from the expert policy, which allows the navigation policy to learn on accurate samples in the dataset.

\section{Multi-Sensory Semi-Supervised Autonomous Mobile Robotic Navigation}\label{sec:design}
We consider a mobile robot navigating in indoor environments of pedestrians and obstacles. The robot's goal is to navigate rapidly and safely in indoor environments. Our goal is to minimize human supervision and allow the robot to learn from its own experience. We do not assume the robot has access to a multi-stage motion planer as a reference policy as in \cite{pfeiffer2016perception}. 

We propose and implement a system, \textit{\textbf{M}ulti-\textbf{S}ensory \textbf{S}emi-\textbf{S}upervised \textbf{L}earning (MS3L)}, to enable the autonomous robotic navigation. Our system combines four policies $\pi_{h}$, $\pi_{s}$, $\pi_{\theta_{r}}$, and $\pi_{\theta_{n}}$, which respectively represent human policy, sensor policy, recording policy, and navigation policy. By initializing $\pi_{\theta_{n}}$ with $\pi_h$ and constraining learning from a suboptimal policy $\pi_{s}$ with a recording policy $\pi_{\theta_{r}}$, the robot is able to surpass the suboptimal policy and achieve near human performance in a self-supervised manner. 

In this section, we present the detail design and framework of {\it MS3L}, including the robotic platform, the policies, and the training protocol. 
\begin{figure*}[t]
	\centering
	\vspace{-0.3in}
	\includegraphics[scale=0.4]{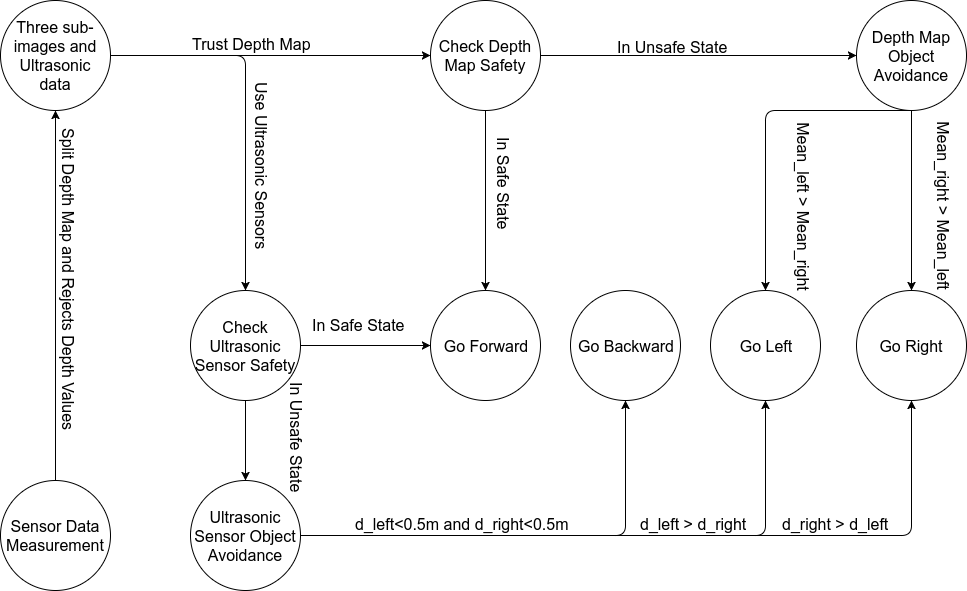}
	\caption{Finite state machine for sensor policy.}
	\label{fig:fsm_sensor}
\end{figure*}

\subsection{Robotic Platform}
\begin{figure}[b]
\center
\includegraphics[scale=0.06]{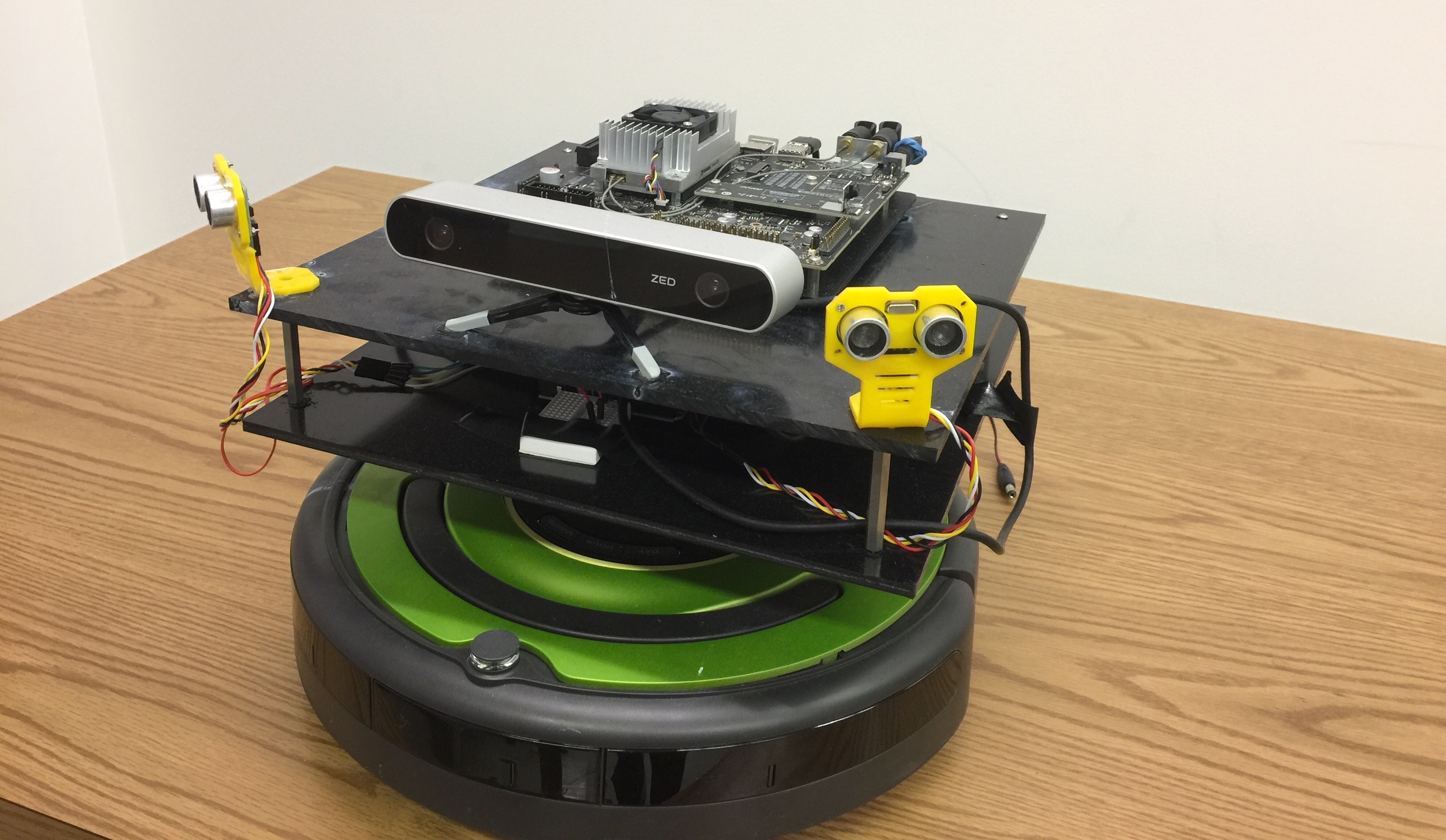}
\caption{iRobot Create2 equipped with a stereo camera, two ultrasonic sensors, and a Jetson TX1.}\label{fig:robot}
\end{figure}

We have built our own robot based on commodity supplies as shown in Fig. \ref{fig:robot}.  The base of the robot is an iRobot Create2\footnote{http://www.irobot.com/About-iRobot/STEM/Create-2.aspx}, which has a linear velocity of [-0.5$m/s$, 0.5$m/s$] and an angular velocity of [-4.5$rad/s$, 4.5$rad/s$]. The robot is equipped with two ultrasonic sensors for distance measurements and a ZED\textsuperscript{\tiny\texttrademark} Stereo Camera\footnote{https://www.stereolabs.com/} to capture RGB images and as well as depth images.  The ultrasonic sensors can detect objects within a distance of [5$cm$, 400$cm$] (or [2$in$, 156$in$]). The valid depth estimation of the stereo camera falls into [0.5$m$, 20$m$]. These two types of depth/distance sensors, namely ultrasonic and stereo camera, complement with each other to derive our sensor policy. In addition, a NVIDIA\textsuperscript{\tiny\textregistered} Jetson TX1\footnote{http://www.nvidia.com/object/embedded-systems-dev-kits-modules.html}  is used as the robot's brain to make inferences, which has 256 CUDA cores with 2GB memory to support a moderate-sized neural network in real-time computation. 

\subsection{\textit{MS3L} Policies}
The {\it MS3L} solution has four policies including {\it human policy ($\pi_h$), sensor policy ($\pi_s$), navigation policy ($\pi_{\theta_n}$)} and {\it recording policy ($\pi_{\theta_r}$)}, designed for the multi-sensory imitation learning. Among those, {\it human policy} and {\it sensor policy} are also referred as {\it reference} policies ($\pi_{ref}$) to guide the navigation.

\subsubsection{Human Policy}
Human policy $\pi_h$ refers to a human operation that guides the robot to navigate through environments and avoid collision. The operator provides initial demonstrations using a joystick to tele-operate the robot remotely. Note that the human policy is used only ONCE in data collection for each of the navigation and recording policy.

\subsubsection{Sensor Policy} Sensor policy $\pi_s$ is used in the self-supervised learning phase to generate suboptimal action labels. This policy generates a set of basic robot control commands: turning, acceleration, and deceleration. Unlike in simulated or controlled environments where dynamics are always known, it is not possible to obtain an accurate model of an unstructured real indoor environment. Our goal is not to use an expensive accurate range sensor like a laser device used in \cite{pfeiffer2016perception} to create an optimal depth control policy, but to build a suboptimal sensor policy $\pi_s$ upon the coarse measurements of those unreliable cheap commodity sensors. 

Sensor policy $\pi_s$ is high bias and low variance due to the hand engineered control and noisy measurements in sensor data. It is regulated through a finite state machine as shown in \textbf{Fig. \ref{fig:fsm_sensor}}. It takes the current depth map $M$ and ultrasonic sensor measurements $d_{left}$ and $d_{right}$ as inputs and generates a control command $a = \pi_{s}(M, d_{left}, d_{right})$. At the first stage, it splits the depth map $M$ into three sub-image. %and calculate mean depth value of each sub-image. These values estimate the object occupancy in left, middle, and right directions. 
Then, it rejects depth values in each sub-image which has the value larger than two standard deviation of the entire depth image %with the standard %those map data, comment-1.2 %deviation $>2\sigma$ ($\sigma$ is the standard deviation of the entire depth map) from the mean of entire depth values are rejected. 
This checks whether we can trust the current depth estimation. If the number of rejected data is large, the robot distrusts the depth map, and rather uses the two ultrasonic sensor measurements ($d_1$, $d_2$) to calculate robot control actions. This usually happens when the robot is facing towards a plain wall or objects only appear in one camera. If the depth image is used, mean depth value of the middle sub-image is used, e.g. the value is smaller than 0.8m in our experiments, to determine whether the robot needs to take object avoidance actions. If the ultrasonic sensor is used, $d_{left}$ and $d_{right}$ are used to check whether the robot is in the safety position. Go back action is only taken when ultrasonic sensor is used because ZED stereo camera is not valid in detecting distance within 0.5 meters.
 %and control commands are taken based on the comparison of these values as shown in the finite state machine graph.

It should be noted that, $\pi_{s}$ is not robust and we expect the recording policy described later to constrain the robot to only learn a subset of data generated by $\pi_{s}$
% It should be noted that, because of the high bias in this algorithm, the sensor policy $\pi_s$ alone is not reliable. Therefore, it needs to be constrained by a recording policy described later to train the robot's navigation policy. 
%it is combined with a recording policy described later to achieve high reliability in use. 

\subsubsection{Navigation Policy}
\begin{figure*}[t]
	\centering
	\includegraphics[scale=0.4]{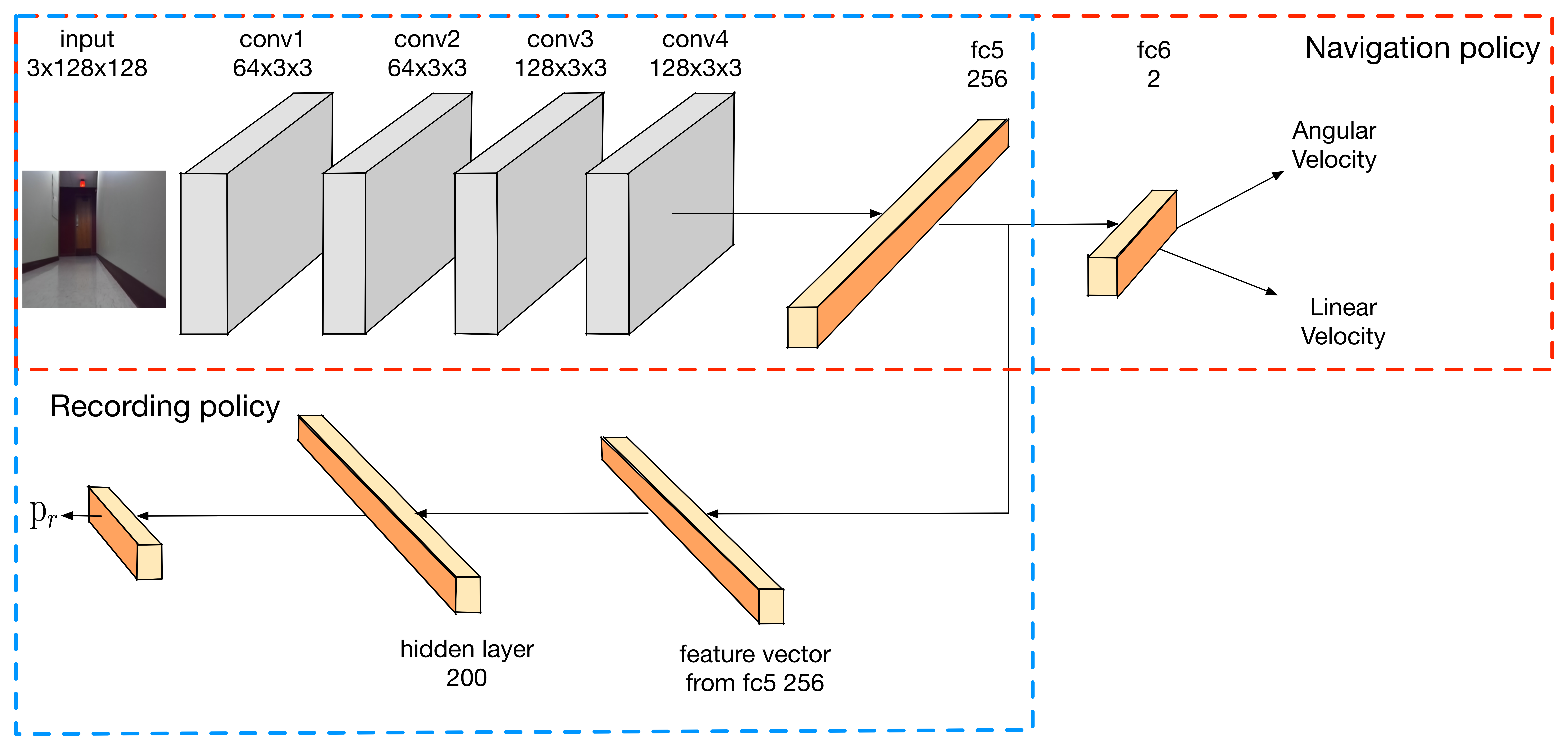}
	\vspace{-0.1in}
	\caption{Our model consists of two policies: navigation policy $\pi_{\theta_n}$ and recording policies $\pi_{\theta_r}$}\label{fig:network}
	\vspace{-0.2in}
\end{figure*}

Navigation policy $\pi_{\theta_n}$ outputs an action to avoid obstacles based only on the current RGB image observed from ZED stereo camera. To handle high dimensional observations i.e. images in our case, we parameterize $\theta_n$ as a 5-layer convolutional neural network (CNN) as shown in Fig.\ref{fig:network}. We adopt VGG-like architecture \cite{simonyan2014very} in our design. All convolution layers have a $3\times3$ kernel size. The first two convolution layers have 64 channels followed by a max pooling layer. The last two convolution layers have 128 channels followed by a fully-connected layer with 256 neurons. ReLU activation functions are applied to all layers. We normalize iRobot Create2 linear and angular velocities between $[-1, 1]$ when collecting dataset. To bound with the this normalized scale, we add a $tanh$ activation function before the output. The policy outputs 2-dimensional actions in forms of linear velocity and angular velocity. Input images are resized to $128\times128$ for real-time inference.

We define the navigation policy dataset as $D_{\pi_{\theta_n}} = \{(x_1, \pi_{ref}(x_1)),...,(x_i, \pi_{ref}(x_i)),...,(x_t,\pi_{ref}(x_t))\}$, where $x_i$ represents the observation at timestep $t$, and $\pi_{ref}(x_i)$ refers to the output of reference policies consisting of human policy $\pi_{h}$ that is used only in the first iteration and sensor policy $\pi_{s}$ that is used in the rest of training iterations. The imitation objective is defined as
\begin{equation}\label{imitation}
C_{i}(D_{\pi_{\theta_n}}, \pi_{{\theta}_n}) =1/N \times \sum_{i=1}^{i=N} {||\pi_{{\theta}_n}(x_i) - \pi_{ref}(x_i)||}_2^2,
\end{equation}
where $N$ is the mini batch size for training. With this objective function, the navigation policy is trained to imitate from both human and sensor policies: $\pi_{h}$ and $\pi_{s}$. 

\subsubsection{Recording Policy}
Recording policy $\pi_{{\theta}_r}$ is crucial to our learning framework, where $\theta_r$ is parameterized by a 2-layer fully-connected neural network. It takes a feature vector from the last fully-connected layer $fc5 = \pi_{{\theta_n}}^5(x)$ of the navigation policy as input and generates the probability $p_r = \pi_{\theta_r}(fc5)$ that the current observation is needed for the navigation policy to learn, as shown in Fig.\ref{fig:network}, where $x$ is observation and ${\theta_n^5}$ represents the parameters of last fully-connected layer of the navigation policy. This design choice of using shared convolutional layers for navigation and recording policies is based on two factors. First, the limited GPU memory on Jetson TX1 is not powerful enough for two CNNs: one for the navigation policy and the other for recording policy. Second, the feature extracted by the CNN at layer ${fc}_5$ can be shared and reused by both the navigation policy and the recording policy to improve its learning speed \cite{Lecun2015Deep}. 

We use the sensor policy $\pi_{s}$ as a suboptimal policy to self label data after the human operated pre-training stage. In order to throttle the propagation of learning errors from $\pi_{s}$, the recording policy only keeps those labelled data from $\pi_s$ if $\pi_{\theta_n}$ deviates far from both reference policies. We define the deviation as the squared difference between the output of the navigation policy $\pi_{{\theta}_n}$ and that of reference policies ($\pi_{h} \text{ and }\pi_{s}$) as
\begin{equation} \label{eq:r_error}
\begin{split}
e(x,\pi_{{\theta}_n}, \pi_{h}, \pi_{s}) = \gamma  ||\pi_{{\theta}_n}(x) - \pi_{h}(x)||_2^2 \\+ (1-\gamma)||\pi_{{\theta}_n}(x) - \pi_{s}(x)||_2^2,
\end{split}
\end{equation}
where $\gamma$ is a constant between 0 and 1 that weights the relative importance of the two reference policies, e.g. if $\gamma$ is closer to 1, the sensor policy is less accounted for the deviation. With this deviation metric, a binary function indicating whether to record is defined as
\begin{equation} \label{eq:r_binary}
\epsilon(x, \pi_{{\theta}_n}) = 
	\begin{cases}
	1, & \text{if } e(x,\pi_{{\theta}_n}, \pi_{h}, \pi_{s}) > \tau \\
	0, & \text{otherwise}
	\end{cases},
\end{equation}
where $\tau$ is a scalar indicating the degree of error tolerated by the recording policy. 

% In our experiment, we set it to a large value 0.8.

$\pi_{{\theta}_r}$ is trained on a recording dataset $D_{{\pi}_{{\theta}_r}}=\{\pi_{{\theta_n}}^5(x_1), ..., \pi_{{\theta_n}}^5(x_n)\}$, which is collected using extra human demonstrations, from the last fully-connected layer $fc5$. The objective function to be minimized by $\pi_{{\theta}_r}$ is a binary cross-entropy loss function defined as
\begin{equation*}\label{eq:cross_entropy}
\hspace{-0.1in}
\begin{multlined}
C_{r}(D_{{\pi}_{{\theta}_r}},\epsilon, \pi_{{\theta}_r} \pi_{{\theta}_n},\pi_{ref})=
-\frac{1}{N}\times  \\ {\sum_{n=1}^N{\epsilon(x_n, \pi_{{\theta}_n}) log(\pi_{{\theta}_r}(x_n)) 
		+(1-\epsilon(x_n, \pi_{{\theta}_n}))log(1-\pi_{{\theta}_r}(x_n))}},
\end{multlined}
\end{equation*}
where $\pi_{ref}$ is the reference policies consisting of $\pi_{h}$ and $\pi_{s}$.
%aling horizontally.
\begin{figure}[h]
	\begin{subfigure}{.24\textwidth}
	%\centering
	\hspace{-0.2in}
	\includegraphics[scale=0.25]{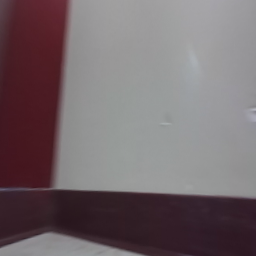}%
	\includegraphics[scale=0.25]{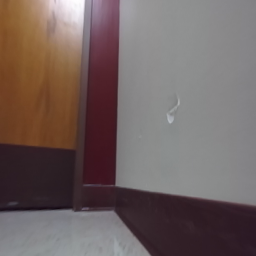}
	\caption{$\beta=0.99$}
  	\label{fig:tau_0.99}
	\end{subfigure}
	\begin{subfigure}{.24\textwidth}
	\centering
	%\hspace{-0.1in}
	\includegraphics[scale=0.25]{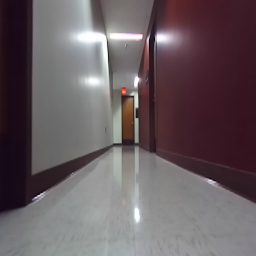}%
	\includegraphics[scale=0.25]{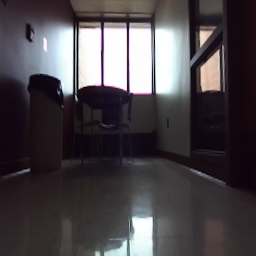}
	\caption{$\beta=0.5$}
  	\label{fig:tau_0.5}
	\end{subfigure}
\caption{Images are uniformly sampled from datasets that have different $\beta$ values.}
\label{fig:tau}
\end{figure}

After the pre-training stage, observations are recorded only if $\pi_{{\theta}_r}(x) > \beta$, where $\beta$ is a recording threshold controlling the degree to which the robot trusts $\pi_{{\theta}_r}$. We refer the observations recorded by $\pi_{{\theta}_r}$ as hard samples. Some examples with different thresholds are shown in Fig.\ref{fig:tau}. Fig.\ref{fig:tau_0.99} shows the images recorded in dangerous states i.e. turning or being close to walls. In contrast, Fig. \ref{fig:tau_0.5} shows the images of going forward, which are not necessary for the navigation policy to learn. A larger $\beta$ places more constraints to the number of observations recorded and labeled by $\pi_s$.

\subsection{Training Protocol}

\begin{algorithm}[ht]

\caption{Multi-Sensory Self-Supervised Learning}\label{alg:mssl}

\begin{algorithmic}[h]
\Procedure{ \textit{MS3L} Training Protocol}{}
\State Randomly initialize ${\pi}_{{\theta}_r}$ and ${\pi}_{{\theta}_n}$.

\State \textbf{Pre-training}

\State Collect $D_{{\pi}_{{\theta}_r}} $ and $D_{{\pi}_{{\theta}_n}}^0$ using $\pi_{h}$

\State ${\pi}_{{\theta}_n^0} = {arg\,min}_{{\theta}_n} C_{i}(D_{{\pi}_{{\theta}_n}}^0, \pi_{{\theta}_n})$

\State ${\pi}_{{\theta}_r^0} = {arg\,min}_{{\theta}_r} C_{r}(D_{{\pi}_{{\theta}_r}} \cup D_{{\pi}_{{\theta}_n}}^0,\epsilon, \pi_{{\theta}_r}, \pi_{{\theta}_n^0},\pi_{ref})$

\State

\State \textbf{Self-supervised learning}

\For{$i=1 \text{ \textbf{to} } k$}
	\State Execute ${\pi}_{{\theta}_n^{i-1}}$, collect and label observations into $D$ using $\pi_{s}$ and $\pi_{\theta_r^{i-1}}$.
	
	\State Only keep $(x, \pi_{ref}(x_i))$ pair from $D$ if $\pi_{{\theta}_r^i}(\pi_{\theta_n^{i-1}}^5(x)) > \beta$
	
	\State $D_{{\pi}_{{\theta}_n}}^i = D \cup D_{{\pi}_{{\theta}_n}}^{i-1}$ 
	
	\State ${\pi}_{{\theta}_n^i} = {arg\,min}_{{\theta}_n} C_{i}(D_{{\pi}_{{\theta}_n}}^{i-1}, \pi_{{\theta}_n^{i-1}})$
	
	\State ${\pi}_{{\theta}_r^i} =  {arg\,min}_{{\theta}_r} $
	\State \indent $C_{r}(D_{{\pi}_{{\theta}_r}} \cup D_{{\pi}_{{\theta}	
	_p}}^i,\epsilon, \pi_{{\theta}_r^{i-1}}, \pi_{{\theta}_n^{i-1}},\pi_{ref})$
\EndFor

\Return{${\pi}_{{\theta}_n^4}$ and ${\pi}_{{\theta}_r^4}$}
\EndProcedure
\end{algorithmic}
\end{algorithm}
With these policies, the training procedure of {\it MS3L} is performed as: the navigation policy $\pi_{\theta_n}$ is initialized by human policy $\pi_h$ and iteratively trained by a suboptimal internal policy $\pi_s$ constrained on a recording policy $\pi_{\theta_r}$.
Our training procedure consists of two stages with five iterations in total, which is shown in Algorithm \ref{alg:mssl}. The first stage is a pre-training stage, where a human operator controls the robot using a PlayStation wireless controller to navigate through the environments and collect the initial recording and navigation policies datasets: $D_{{\pi}_{{\theta}_r}}$ and $D_{{\pi}_{{\theta}_n}}^0$ to initialize the navigation policy: $\pi_{\theta_r}$ and $\pi_{\theta_n}$. These two datasets are collected in opposite directions shown in Fig.\ref{fig:floor}. to ensure that the recording policy correctly classifies hard observations. While the navigation policy is trained to have a basic understanding of the environment, it does not learn any complex actions such as turning or decelerating. A self-supervised learning stage is proposed to further improve the policy without human operators. At this stage, $\pi_{s}$ is used to generate labels for the collected data. In addition, $\pi_{{\theta}_r}$ is used to constrain the data collection to remove largely deviated data. Specifically, an observation $x$ and its corresponding label $\pi_{s}(M, d_1, d_2)$ are only recorded when $\pi_{{\theta}_r}(\pi_{\theta_n}^5(x) )> \beta$. This ensures that the deficient labeling resulted from noisy measurements of $\pi_{s}$ is minimized. Then, the navigation policy is trained upon the aggregation of the initial dataset and the selected observations. The trained navigation policy and the aggregation of recording and navigation datasets are used to update the recording policy.  

\section{Performance Evaluation}\label{sec:eval}
\begin{figure}[h]
%\hspace{0.62in}
\begin{subfigure}[t]{.2\textwidth}
\centering
\includegraphics[scale=0.034]{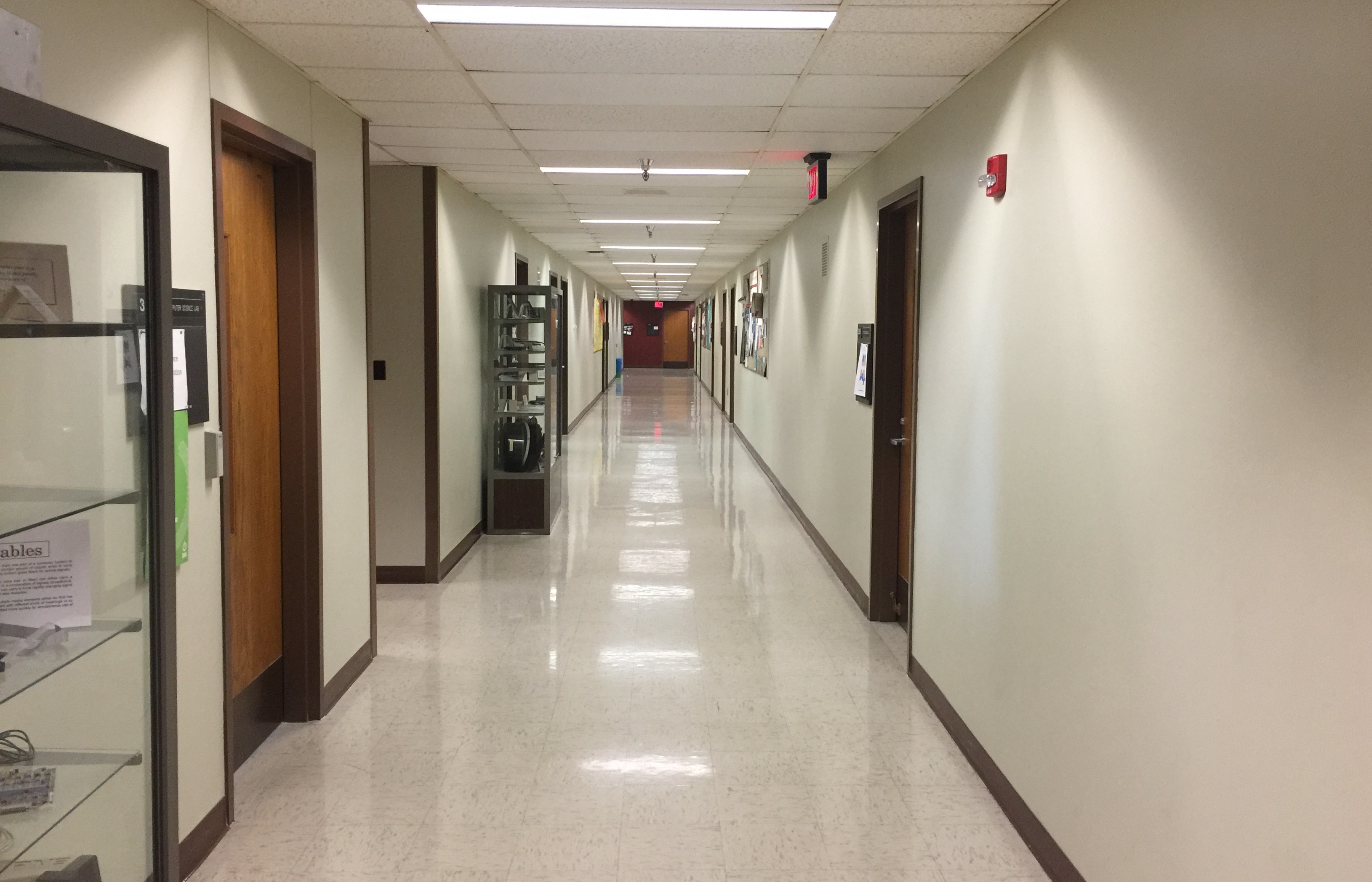}
\caption{Hallway}\label{fig:envs-1}
\end{subfigure}%
\begin{subfigure}[t]{.3\textwidth}
\centering
\includegraphics[scale=0.098]{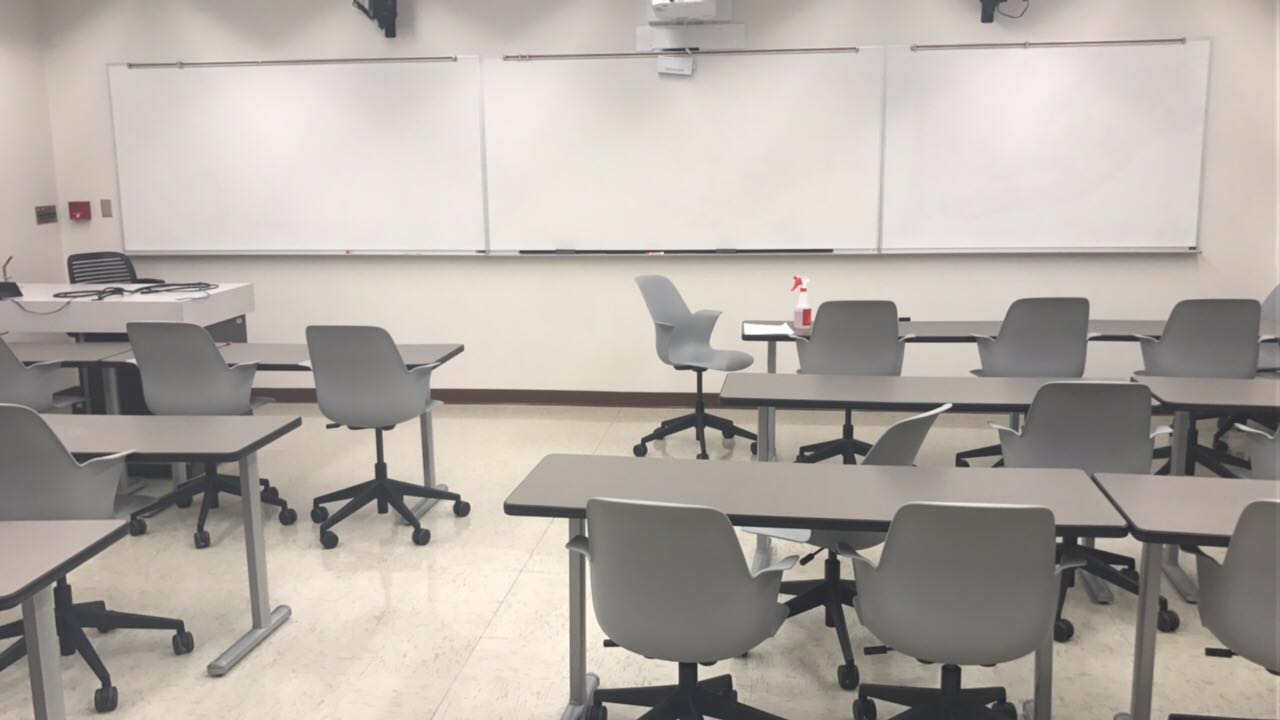}
\caption{Classroom}
\label{fig:envs-2}
\end{subfigure}
\caption{Training environments for robots.}
\label{fig:env}
\end{figure}

\begin{figure*}[t]
\centering
\vspace{-0.2in}
\includegraphics[scale=0.45]{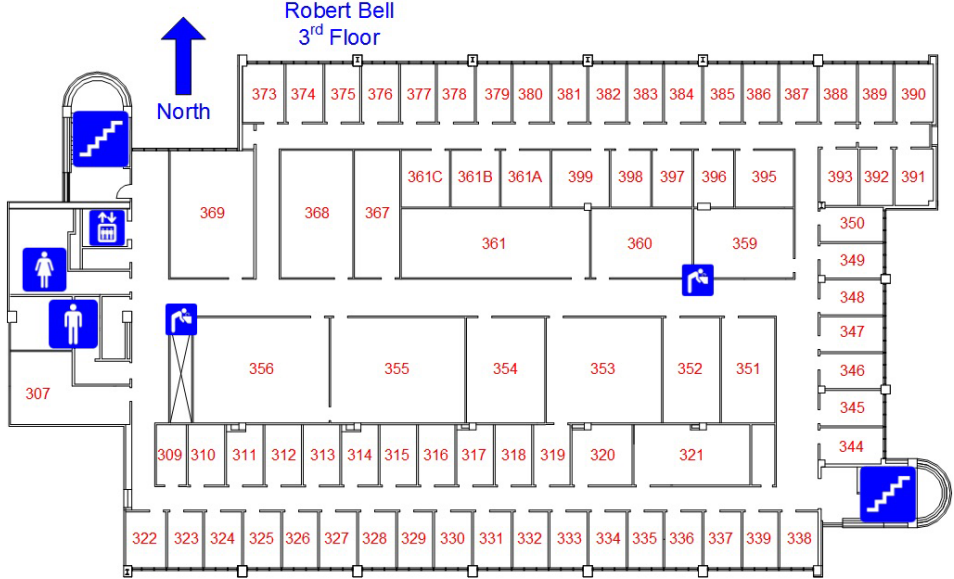}
\vspace{-0.3in}
\caption{Data collection trajectories. Blue arrow indicates the trajectory for collecting recording policy 
	dataset and the red one shows the trajectory for collecting navigation policy datasets for all iterations.}
\label{fig:floor}
\end{figure*}

The experiments have been performed in the 3rd floor of the Robert Bell (RB) Hall building at Ball State University. Fig. \ref{fig:envs-1} and \ref{fig:envs-2} are indoor environments where our robot has been trained and Fig.\ref{fig:floor} shows the floor plan.

We have implemented {\it MS3L} on tensorflow\cite{abadi2016tensorflow}. Adam optimizer\cite{Kingma2014Adam} is used and configured with a learning rate of 0.0001 for $\pi_{{\theta}_n}$ and 0.001 for $\pi_{{\theta}_r}$. Both networks are trained over 50 epochs with a $L2$ weight decay of 0.0001. We set $\gamma = 0.8$ and $\tau=0.00025$ while training $\pi_{{\theta}_r}$. In addition, we set $\beta=0.99$ in data collection using the recording policy. Input images are rescaled to $128 \times 128$ RGB images. We set $k=4$ during self-supervised training. For each training iteration, we run the robot for $250s$ and the images are taken at $30fps$. We use the traveled {\it distance} and {\it time} to collision as evaluation metrics in the experiments. We set the maximum travel duration to be $250s$ throughout the experiments.%Therefore, the maximum travel duration is the test time: $250s$. 

We have trained our robot in two indoor environments: RB 3rd floor hallway as shown as red arrow in Fig. \ref{fig:floor} and in RB-356 classroom during break time. The classroom is an extremely difficult environment, where the robot collides into obstacles even with human operator's supervision in some tests. 
\begin{figure}[h]
\begin{tabular}[c]{cc}
  \hspace{-0.25in}
    \begin{subfigure}[c]{0.3\textwidth}
	\includegraphics[scale=0.35]{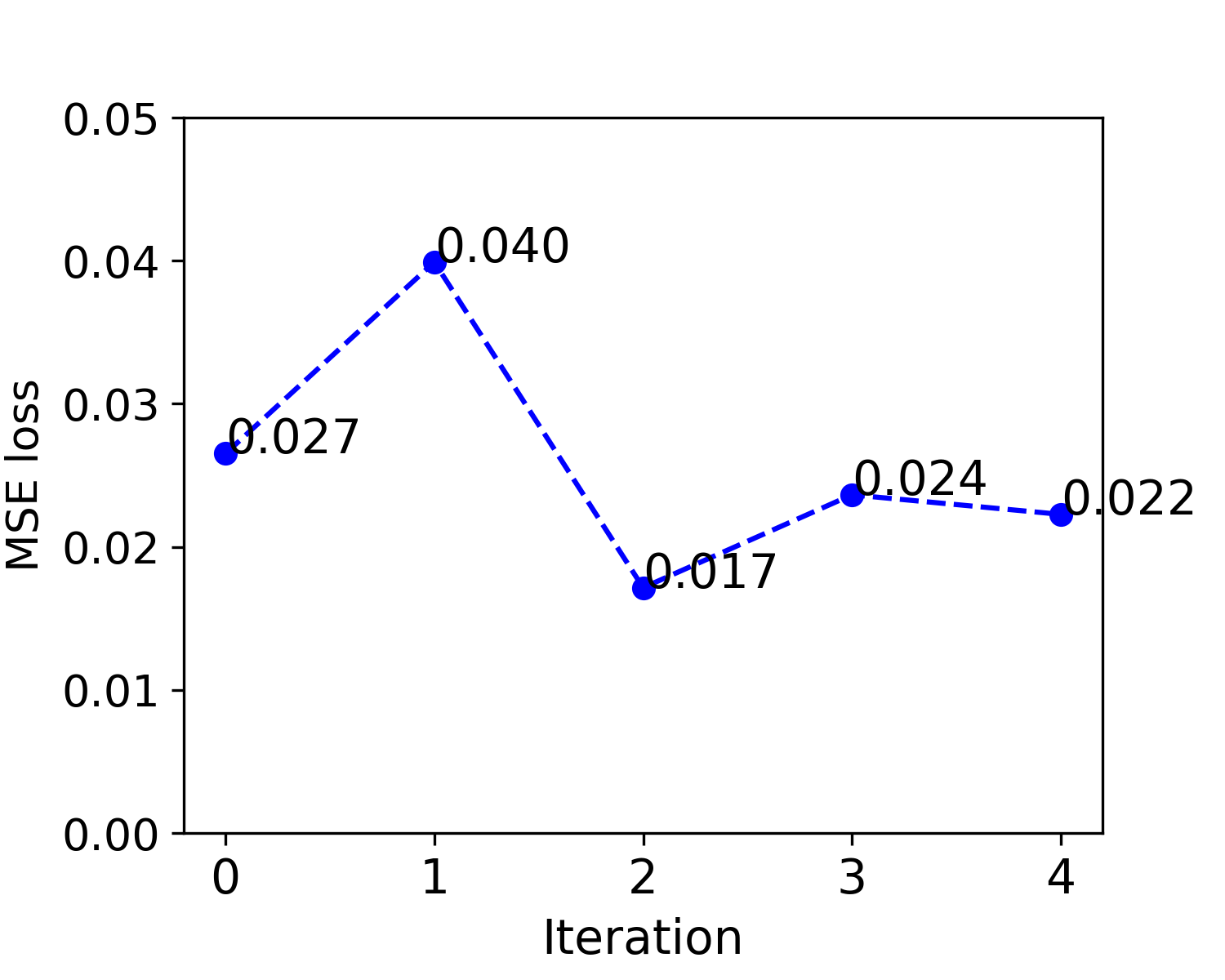}
	\includegraphics[scale=0.35]{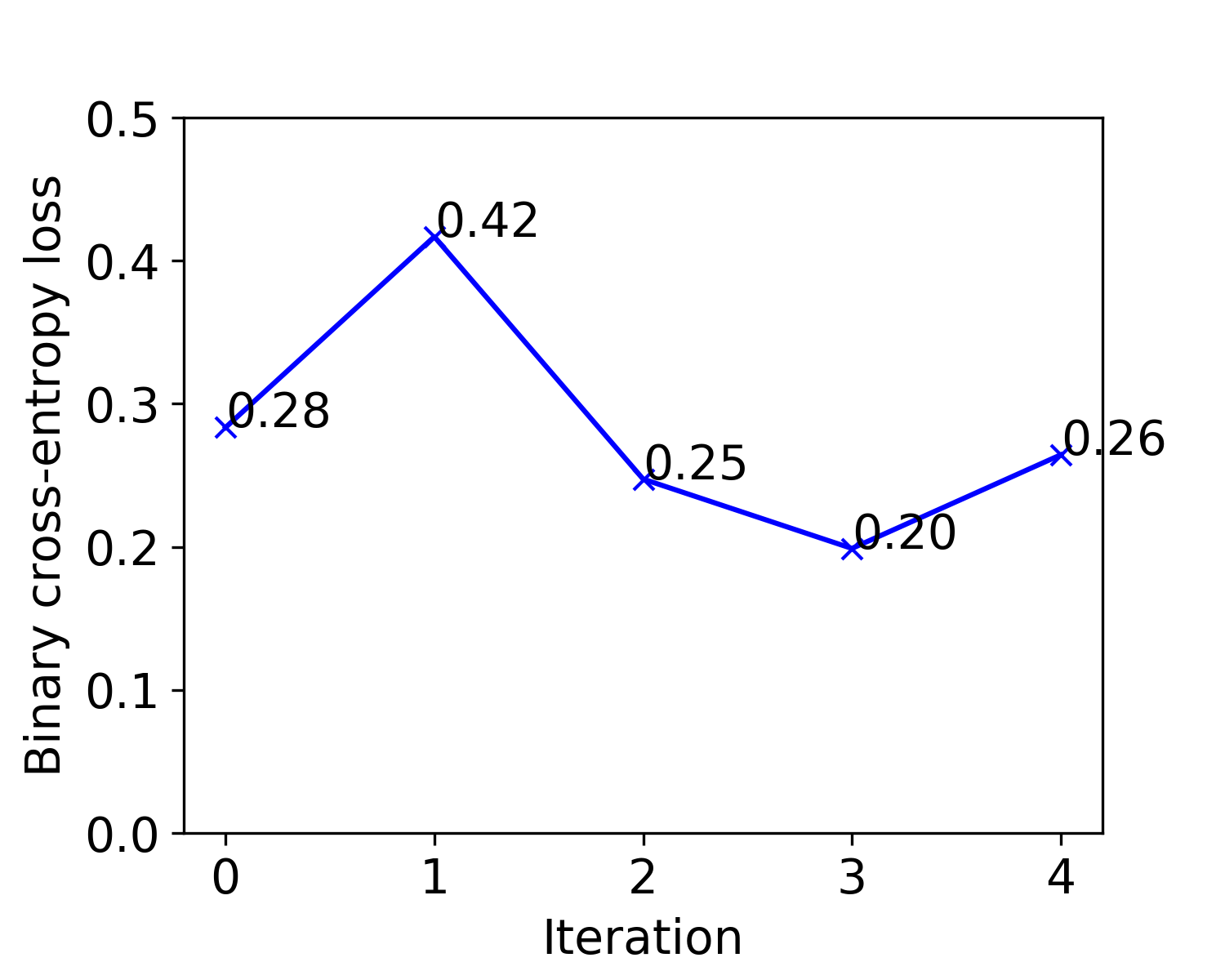}
	\caption{}
	\label{fig:loss}
    \end{subfigure}&
    \hspace{-0.6in}
    \begin{subfigure}[c]{0.3\textwidth}
	\includegraphics[scale=0.35]{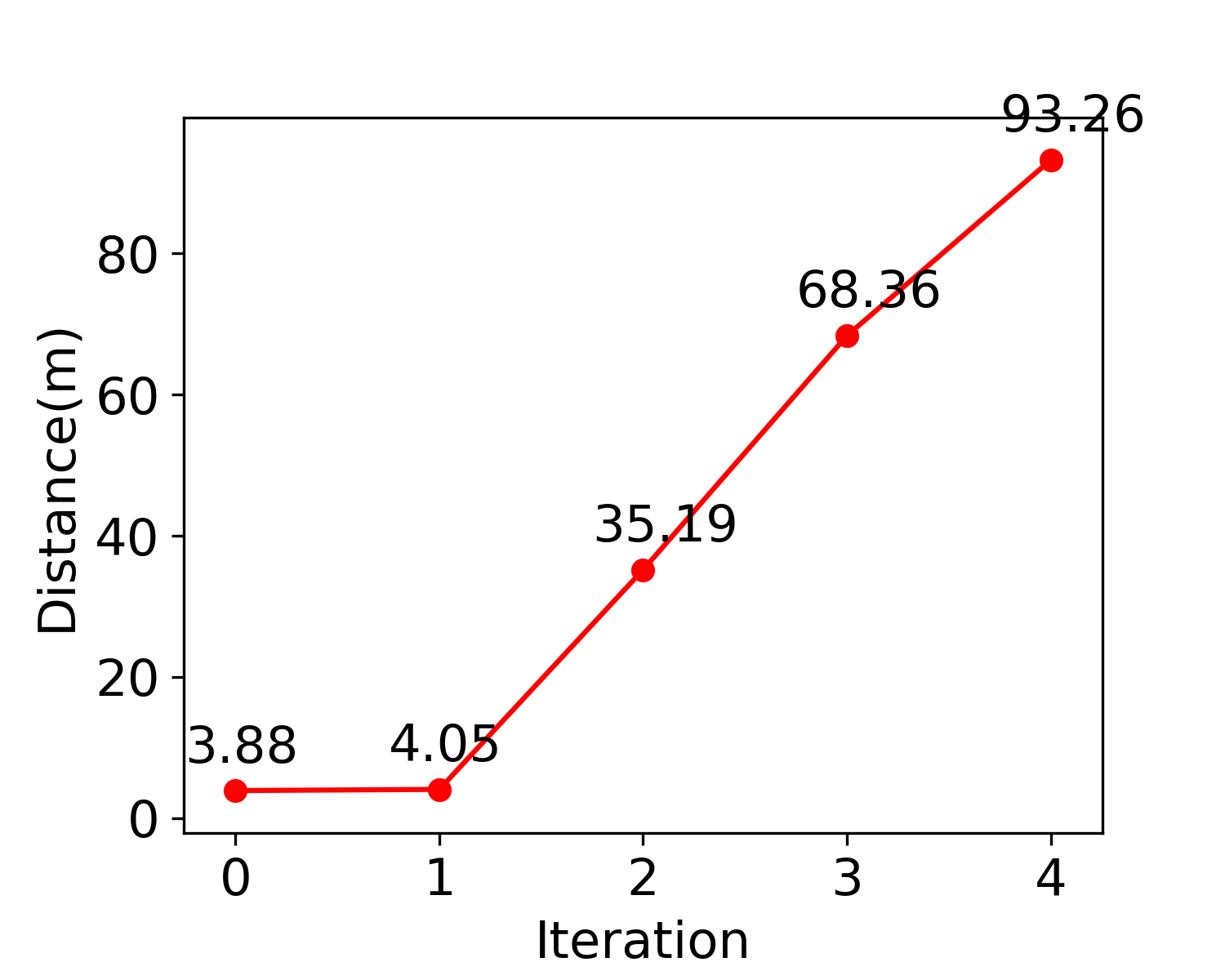}
	\includegraphics[scale=0.35]{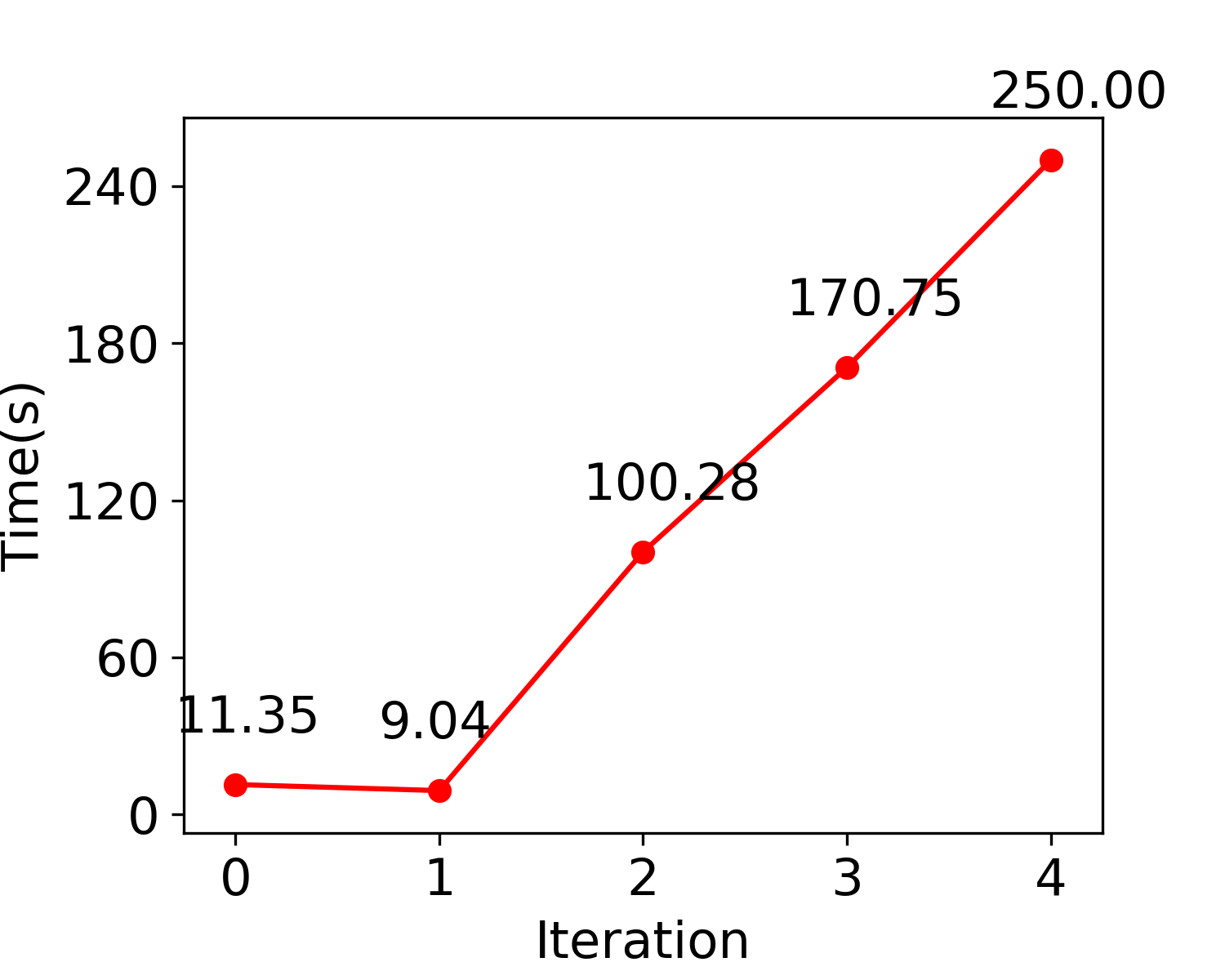}
	\caption{}
	\label{fig:dist_time}
    \end{subfigure}\\
  \end{tabular}    
\caption{Averaged evaluation losses on navigation \textbf{(top)} and recording policies \textbf{(bottom)} over five training iterations are shown in \textbf{(a)}. Traveled distance (\textbf{top}) and time (\textbf{bottom}) to collision in each iteration are shown in \textbf{(b)}.}
\label{fig:evaluation}
  \vspace{-0.2in}
\end{figure}

\subsection{Training Evaluation}

\subsubsection{Validation} We first evaluate the performance of our navigation and recording policies during training process. We have reserved $10\%$ of data samples as the validation set in each iteration. We have measured the losses for navigation and recording policies, which are the mean squared error (MSE) between reference and navigation policies and the binary cross-entropy loss respectively. They are averaged over each training iteration as shown in Fig.\ref{fig:loss}. The averaged losses of navigation and recording policies both reach their largest at the first iteration in the self-supervised stage. We conjecture this is due to the discrepancy between human and sensor policies. After this iteration, the navigation policy rapidly converges to reference policies and then fluctuates within a small range.

\subsubsection{Performance of Navigation Policy} Fig.\ref{fig:dist_time} shows the performance of navigation policy during the five training iterations. It reflects the results of averaged losses in Fig.\ref{fig:loss}: there is no performance gain in the first two iterations and then the performance linearly increases after the first iteration of the self-supervised learning phase. We have noticed that the navigation policy is able to safely navigate the robot across the hallway without any collision within the time limits at the last iteration, which is also clearly indicated by the rightmost point on Fig.\ref{fig:dist_time}. These observations imply that {\it MS3L} can safely and autonomously navigate robots in complex real environments after sufficient iterations of training.

\subsubsection{Performance of Recording Policy} We also analyze the performance of recording policy during each iteration. As an illustration, Table \ref{tb:record} shows the number of training samples collected during each iteration. In the pre-training stage (Iteration $0$), human operator navigates the robot to collect the initial dataset within $250s$ at $30fps$. In the self-supervised learning stage, recording policy is extremely effective in that it drastically reduces the number of collection samples by ruling out those deficient labeled observations. 
\begin{center}
\captionof{table}{Training data collected during 5 iterations}
\label{tb:record}
\centering
{\begin{tabular}{c c c c c c}
\hline
Iteration &{0} &1 & 2 & 3 & 4 \\
\hline
\# of observations & 7500& 351& 853& 790&335\\
\hline
\end{tabular}
}
\end{center}

\begin{figure}[!h]
  \begin{tabular}[c]{cc}
  \hspace{-0.25in}
    \begin{subfigure}[c]{0.28\textwidth}
     	\includegraphics[scale=0.37]{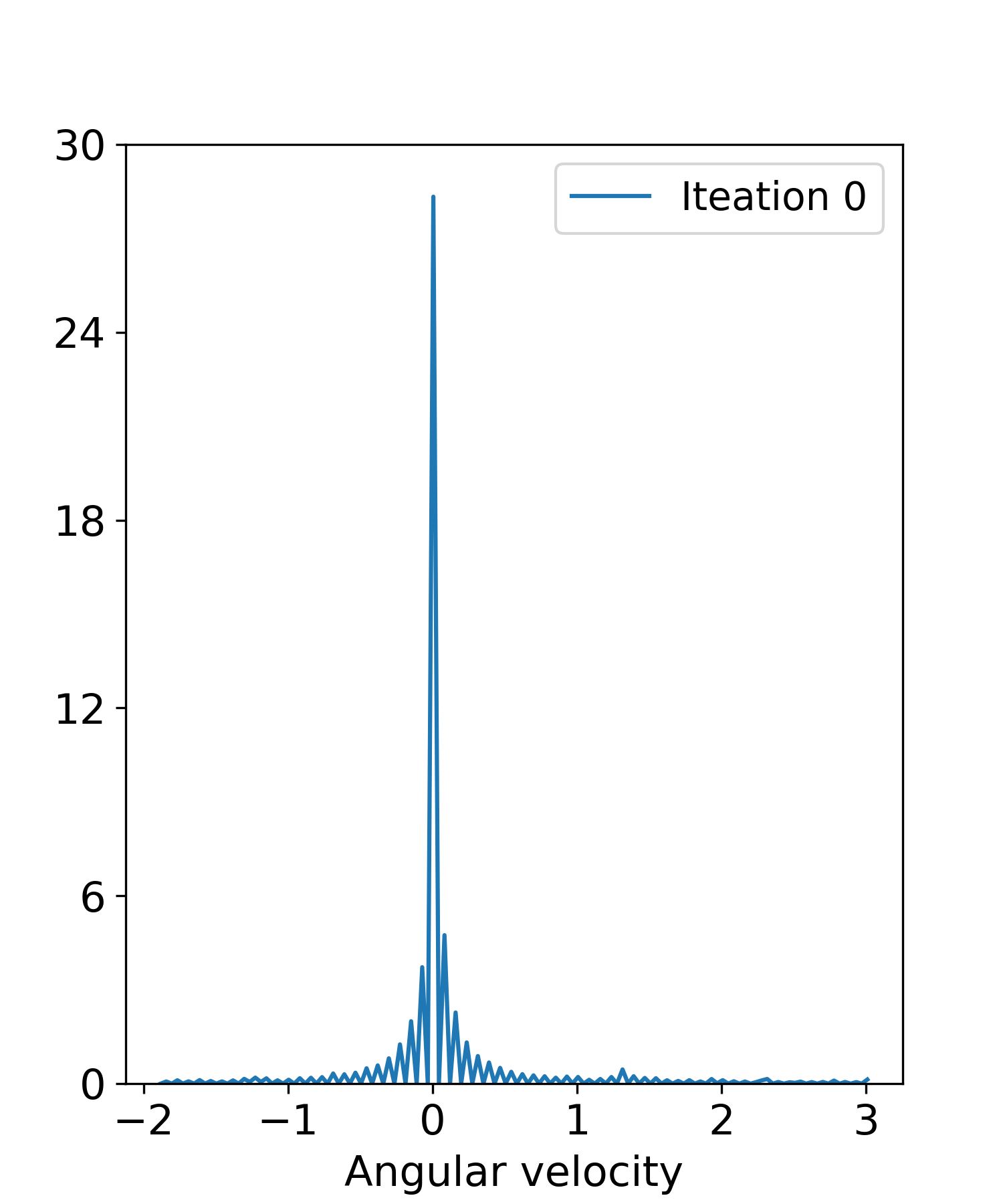}
     	\label{fig:dist}
     	\caption{}
    \end{subfigure}&
    \hspace{-0.3in}
    \begin{subfigure}[c]{0.28\textwidth}
	\includegraphics[scale=0.37]{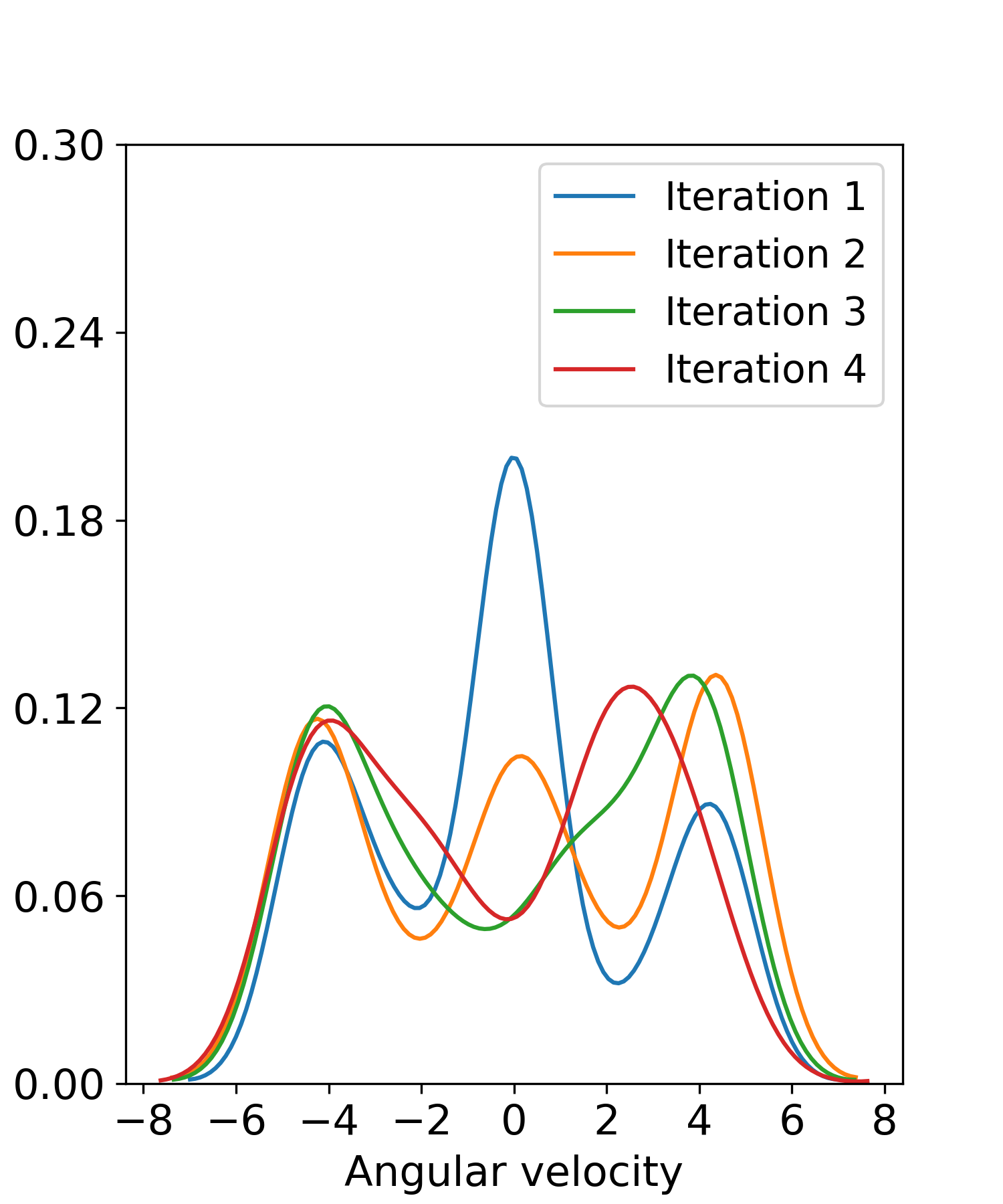}
	\caption{}
	\label{fig:{iter1-4}}
    \end{subfigure}\\
  \end{tabular}    
\caption{Angular velocity distributions over the pre-training stage ({\bf a}) and self-supervised learning stage ({\bf b}).}
\label{fig:distribution}
\vspace{-0.2in}
\end{figure}

%\subsubsection{Performance of Angular Velocity Distribution} 
In addition, distributions for angular velocity during each iteration are are another metric to measure the performance of the recording policy. They are evaluated and presented in Fig.\ref{fig:distribution}. The training samples collected in the pre-training stage primarily consist of data with angular velocity $0rad/s$ as in Fig.\ref{fig:distribution}a, which represents going forward. This singular value can't contribute anything useful to data labeling for training. In contrast, the data collected is more uniformly distributed in self-supervised learning stage as in Fig.\ref{fig:{iter1-4}}. This means these data contain more stateful information such as turning in different angular velocities and are thus significantly useful for training. This also indicates that, in addition to reducing the effect of deficient labeling from sensor policy, {\it MS3L} also ensures only hard samples that have not been seen in the past iterations are recorded and trained by navigation policy. 

\subsection{Recording Threshold $\beta$}
\begin{figure}[h]
	\centering
	\vspace{-0.3in}
	\includegraphics[height=4cm, width=7cm]{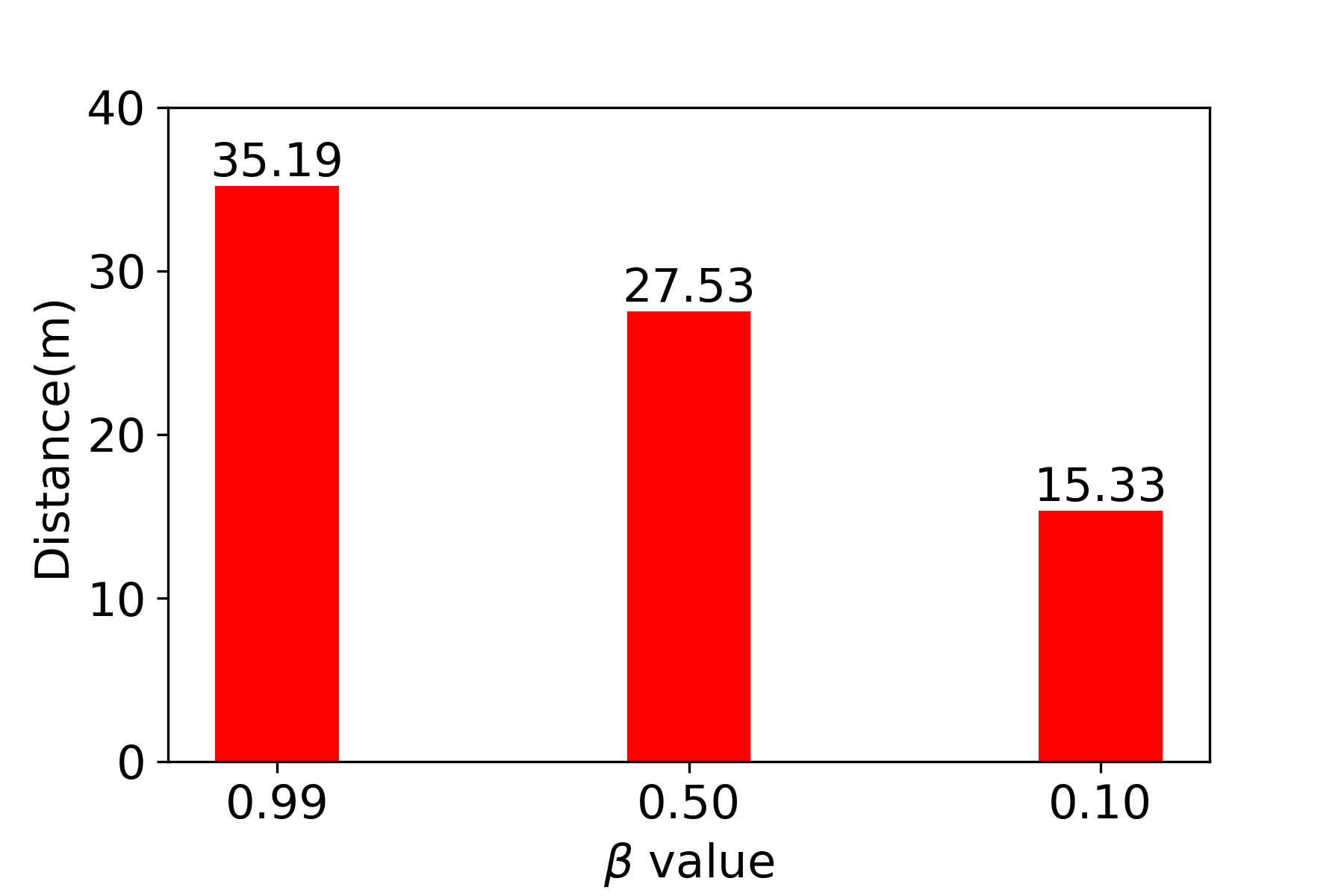}
	\caption{Traveled distances with $\beta$ values of 0.99, 0.5, and 0.1.}
	\label{fig:diff_beta}
	\vspace{-0.2in}
\end{figure}

An important factor that affects the framework and recording policy is threshold $\beta$ that controls what observations should be considered as hard samples. We use three different $\beta$ values to evaluate how it influences the performance of the system as shown in Fig.\ref{fig:diff_beta}. The performance has been evaluated at the third iteration of training process\footnote{Evaluating the performance after the third iteration is dangerous when $\beta$ is 0.5 or 0.1 because the performance of navigation policy degrades quickly.}. As we can observe from the figure, the traveled distance decreases as we reduce $\beta$. We conjecture that this is because the system learns too much from the suboptimal sensor policy with a small $\beta$. When $\beta$ is near 0, the recording policy is merely utilized and the data collection is not constrained, which degrades the performance of navigation policy. A large $\beta$ value is therefore necessary to ensure the recording policy to perform effectively during the self-supervised learning stage.

\subsection{Performance of Navigation After Training: Comparison with Baselines}
After training, it is of utmost interest to assess the actual performance of the trained {\it MS3L} robot in real environments. Since the robot has been trained and learned from human and sensor policies, we define these two policies as baselines. In addition, DAgger algorithm is also used to compare with our framework. The training process of DAgger is the same as our framework except that it requires all $37500$ images to be recorded during the five training iterations and it needs human to correct the actions. It is valuable to compare the robot navigation performance with these baselines. We have tested the comparison over three tasks. For fair comparison, during the tests, the human operator can only perceive the environments at the first-person view from the cameras of the robot as it does at its own. The results show that {\it MS3L} is robust to learn from noisy and suboptimal policy. We have also noticed that {\it MS3L} surpasses the sensor policy by a large margin in most of the tasks, it is also able to achieve near-human performance in two tasks, and even outperforms the human operator in one task. Two human operators participate in each task. Every task is performed 5 times and the results are shown in average distance in meters and time in seconds. The three tasks are presented as follows:

\begin{itemize}
	\item The first task is to navigate the robot through the hallway during normal business time. While our robot is trained in this environment, the training data collected is during break time when there are few people walking in the environment. We set this task during business hours when the walking traffic is heavy inside the hallways to examine how robust {\it MS3L} is to deal with unseen cases after training.
	\item The second task is navigate the robot through a classroom where chairs and tables are main obstacles. This environment is difficult even for human operator.
	\item The third task is navigate the robot through the 3rd floor with Gaussian noise with zero mean and unit standard deviation added to the controller that emulates hardware malfunction. We would like to use this test to show the robustness of {\it MS3L} in case of hardware failures or human manipulation mistakes.
\end{itemize}

\begin{figure}[!h]
\vspace{-0.2in}
\hspace{-0.32in}
  \begin{tabular}[c]{cc}
    \begin{subfigure}[c]{0.53\textwidth}
     	\includegraphics[scale=0.27]{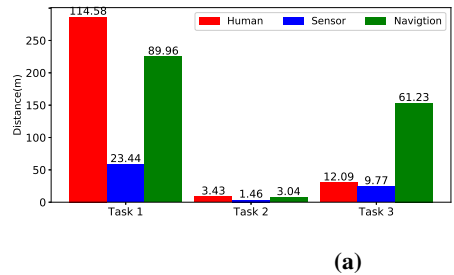}
     	\caption{}
     	\label{fig:distance}
    \end{subfigure}
    \\
    \begin{subfigure}[c]{0.53\textwidth}
	\includegraphics[scale=0.27]{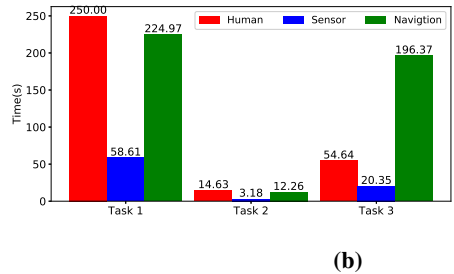}
	\label{fig:duration}
	\caption{}
    \end{subfigure}\\
  \end{tabular}    
\caption{Traveled distance (\textbf{a}) and time to collision (\textbf{b}) in the test stage after training in three tasks.}
\label{fig:baselines}
\vspace{-0.1in}
\end{figure}

Fig.\ref{fig:baselines} shows the results of comparisons in these three tasks. We can observe that the performance of {\it MS3L} exceeds sensor policy in every environment tested. Human policy as the baseline has a slightly higher performance in the first tasks. However, in case of hardware failures or operation mistakes as in the third task, human operator is not able to deal with such unexpected events, while {\it MS3L} is robust to these noises and greatly surpasses human performance. DAgger achieves slightly better performances in all three tasks compared to {\it MS3L}. However, DAgger requires $2.8\times$ more examples to be recorded compared to our framework. In addition, it needs heavy human labeling which does not exist in our work. 

From the experiments, the sensor policy often fails when the robot faces a plain wall or narrow corridor where depth information can not be reliably estimated from stereo images correctly. The navigation policy fails in avoiding multiple objects. This is because the environment is not completely observable and the policy only considers the current observation, not any historical information. For example, it forgets the position of the previous object while trying to avoid the current one. 

\section{Conclusion}\label{sec:conclusion}
In this work, we proposed a solution, Multi-Sensory Self-Supervised Learning(\textit{MS3L}), for autonomous robotic navigation. {\it MS3L} is an imitation deep learning with sensor fusion. It is able to perform robotic navigation tasks in real environments. {\it MS3L} designs a suboptimal sensor policy that replaces human operators after the initial training. A recording policy is then proposed to restrict learning from the suboptimal policy that likely lead to serious robotic damage. Extensive experiments in real indoor environments have demonstrated that {\it MS3L} is able to successfully and reliably surpass the suboptimal policy that it learns from and even outperforms the performance of human operator in unexpected events such as hardware failures.

\section*{Acknowledgment}
The authors would thank all anonymous reviewers for their precious time. This material is based upon work supported by the National Science Foundation under Grant No. \#1408165.

\bibliographystyle{abbrv}
\bibliography{reference.bib}

\begin{IEEEbiography}[{\includegraphics[width=1in,height=1.25in,clip,keepaspectratio]{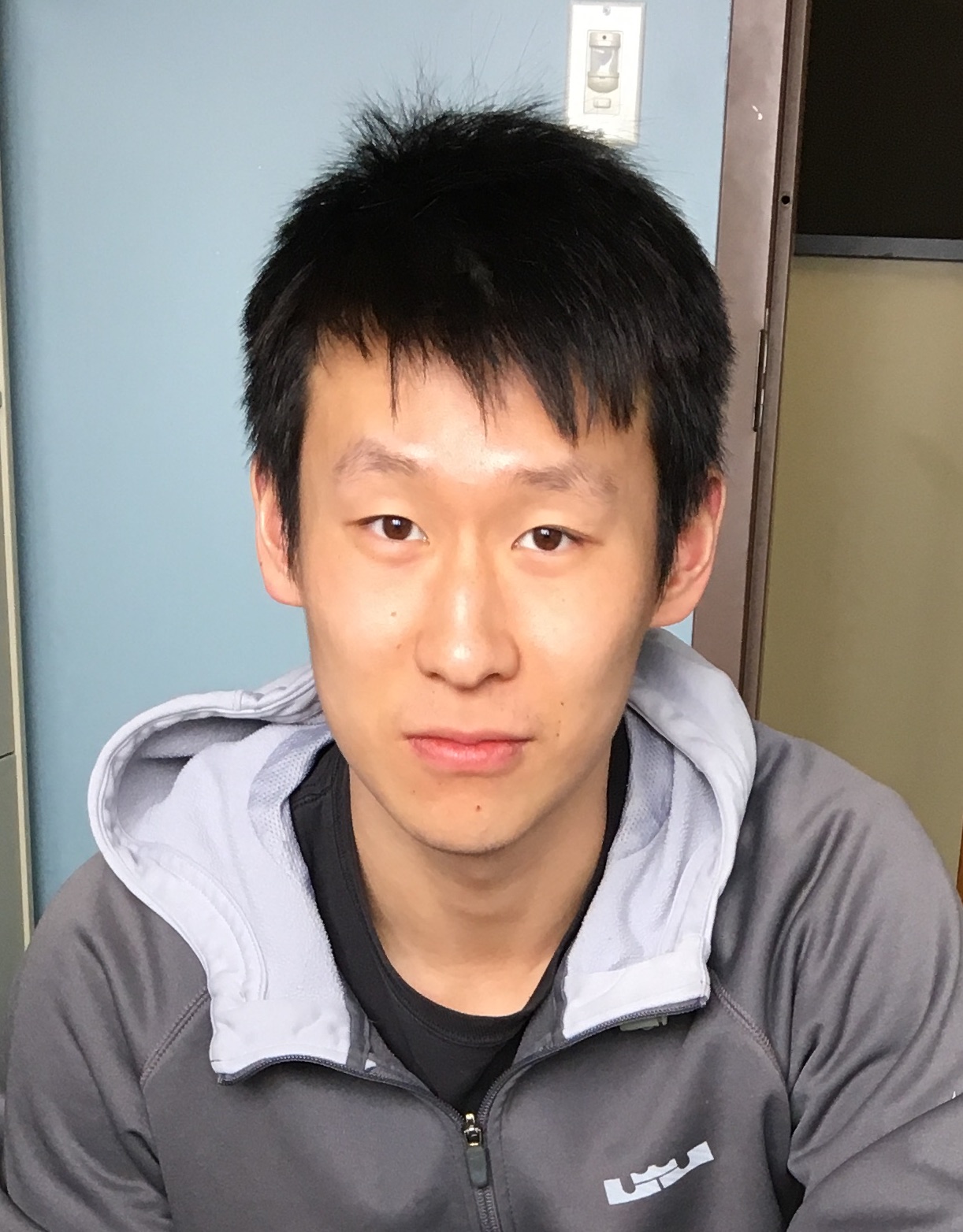}}]{Junhong Xu}
received the B.S. degree in Computer Science from the collaborative program of Ball State University, Muncie, IN, USA and Kunming University of Science and Technology, Kuming, China, in 2016. He is currently pursuing the M.S. degree in Computer Science at Ball State University, Muncie, IN, USA.
\end{IEEEbiography}

\begin{IEEEbiography}[{\includegraphics[width=1in,height=1.25in,clip,keepaspectratio]{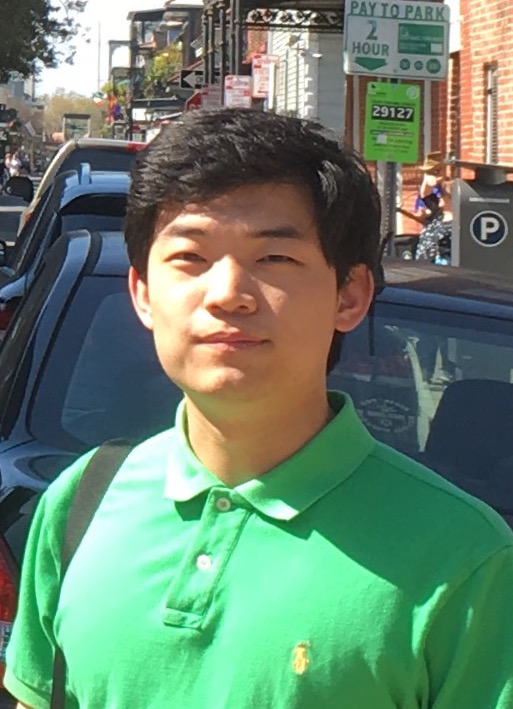}}]{Shangyue Zhu}
received the B.S. degree in Computer Science from the collaborative program of Ball State University, Muncie, IN, USA and  Xi'an University of Posts and Telecommunications, Xi'an, China, in 2015. He received the M.S. degree in Computer Science at Ball State University, Muncie, IN, USA, in 2017.
\end{IEEEbiography}

\begin{IEEEbiography}[{\includegraphics[width=1in,height=1.25in,clip,keepaspectratio]{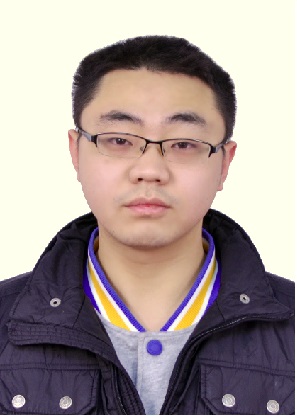}}]{Hanqing Guo}
received the B.S. degree in Telecommunication Engineering from Chongqing University of Posts and Telecommunications, Chongqing, China, in 2015. He is currently pursuing the M.S. degree in Computer Science at Ball State University, Muncie, IN, USA.
\end{IEEEbiography}

\begin{IEEEbiography}[{\includegraphics[width=1in,height=1.25in,clip,keepaspectratio]{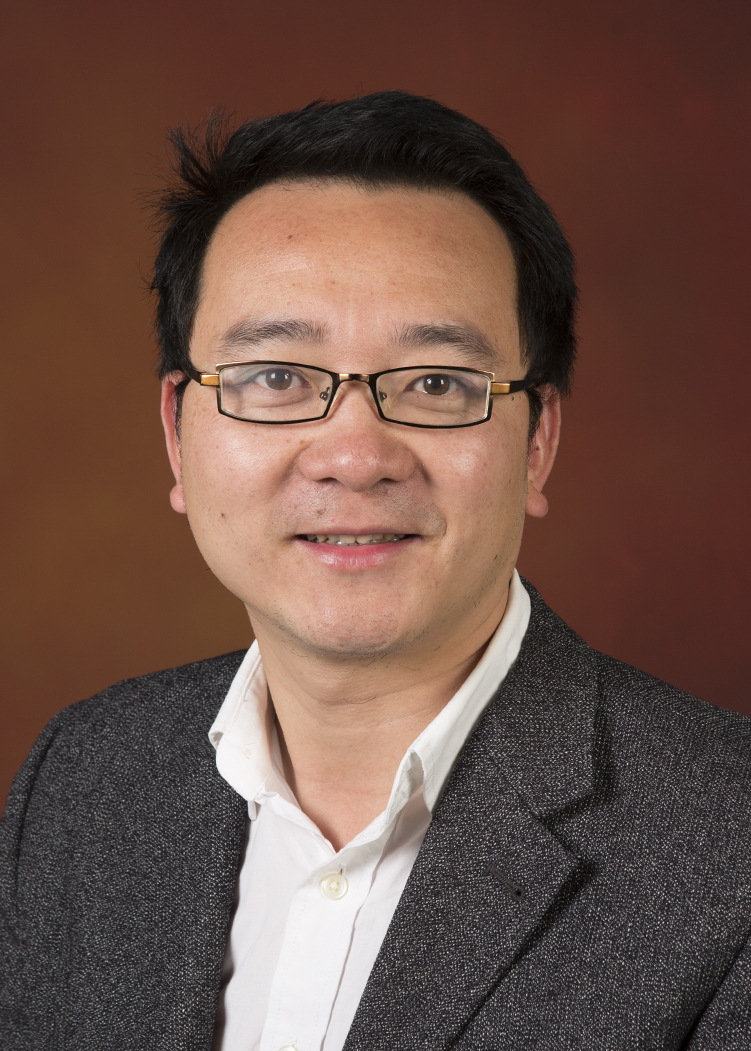}}]{Shaoen Wu}
Shaoen Wu (M'14- SM'16) received the Ph.D. degree in computer science from Auburn University, Auburn, AL, USA, in 2008. 

He is currently an Associate Professor of computer science with Ball State University, Muncie, IN, USA. He was an Assistant Professor with the School of Computing, University of Southern Mississippi, Hattiesburg, MS, USA, a Research Scientist with ADTRAN Inc., Huntsville, AL, USA, and a Senior Software Engineer with Bell Laboratories, Qingdao, China. His current research interests include wireless and mobile networking, cyber security, cyber-physical systems, and cloud computing. 

Prof. Wu has served on the chairs and the committees of various conferences, such as the IEEE ICNC, IEEE INFOCOM, ICC, and Globecom, and an Editor for several journals. He was a recipient of the Best Paper Award of the IEEE ISCC 2008 and the ANSS 2011.  
\end{IEEEbiography}

\end{document}